\pdfoutput=1
\documentclass{article}

\usepackage[preprint]{neurips_2023}

\usepackage[utf8]{inputenc} 
\usepackage[T1]{fontenc}    
\usepackage{hyperref}       
\usepackage{url}            
\usepackage{booktabs}       
\usepackage{amsfonts}       
\usepackage{nicefrac}       
\usepackage{microtype}      
\usepackage[table,xcdraw]{xcolor}         

\usepackage{subcaption}
\usepackage{wrapfig}
\usepackage{multirow}
\usepackage{makecell}
\usepackage{xspace}
\usepackage{enumitem}
\usepackage{amsmath}
\usepackage{listings}
\usepackage{algorithm}
\usepackage{algorithmic}
\usepackage{amsfonts,amssymb}
\usepackage{mathrsfs}
\usepackage{subcaption}
\usepackage{natbib}
\setcitestyle{numbers,square}
\usepackage{graphicx}       
\usepackage{cleveref}
\usepackage{tabularray}
\usepackage{pifont}
\usepackage[most]{tcolorbox}
\usepackage{lipsum}
\usepackage[table]{xcolor}

\newcommand{\hhide}[1]{}
\newcommand{\hide}[1]{}

\newcommand{\model}[0]{\textsc{LogicGame}\xspace}
\newcommand{\cmark}{\textcolor{green}{\ding{51}}} 
\newcommand{\xmark}{\textcolor{red}{\ding{55}}} 
\definecolor{oneDColor}{RGB}{255,240,240}  
\definecolor{twoDColor}{RGB}{240,240,255}  
\newtcolorbox{promptbox}[1][]{
    colback=gray!5,
    colframe=gray!50!black,
    fonttitle=\bfseries,
    title=#1,
    arc=0pt,
    outer arc=0pt,
    boxrule=0.5pt,
    top=6pt,
    bottom=6pt,
    left=6pt,
    right=6pt,
}

\title{\textsc{LogicGame}: Benchmarking Rule-Based Reasoning Abilities of Large Language Models }

 \author{
Jiayi Gui$^{1\dagger*}$ , Yiming Liu$^{2}$\footnotemark[1] , Jiale Cheng$^{12\dagger}$\footnotemark[1] , Xiaotao Gu$^{1}$\footnotemark[1] , Xiao Liu$^{12}$ , \\ \textbf{Hongning Wang$^{2}$ , Yuxiao Dong$^{2}$ , Jie Tang$^{2}$ , Minlie Huang$^{2}$\footnotemark[3]}
\\ \\
$^1$Zhipu.AI \quad $^2$Tsinghua University \\
}

\begin{document}

\maketitle

\thispagestyle{plain}

\renewcommand{\thefootnote}{\fnsymbol{footnote}}
    \footnotetext[1]{JG, YL, JC, and XG contributed equally.}
    \footnotetext[2]{Work done when JG and JC interned at Zhipu AI.}
    \footnotetext[3]{Corresponding author.}
\renewcommand{\thefootnote}{\arabic{footnote}}

\begin{abstract}


Large Language Models (LLMs) have demonstrated notable capabilities across various tasks, showcasing complex problem-solving abilities. Understanding and executing complex rules, along with multi-step planning, are fundamental to logical reasoning and critical for practical LLM agents and decision-making systems. However, evaluating LLMs as effective rule-based executors and planners remains underexplored.
In this paper, we introduce \model, a novel benchmark designed to evaluate the comprehensive rule understanding, execution, and planning capabilities of LLMs. Unlike traditional benchmarks, \model provides diverse games that contain a series of rules with an initial state, requiring models to comprehend and apply predefined regulations to solve problems.
We create simulated scenarios in which models execute or plan operations to achieve specific outcomes. These game scenarios are specifically designed to distinguish logical reasoning from mere knowledge by relying exclusively on predefined rules. This separation allows for a pure assessment of rule-based reasoning capabilities.
The evaluation considers not only final outcomes but also intermediate steps, providing a comprehensive assessment of model performance. Moreover, these intermediate steps are deterministic and can be automatically verified.
\model defines game scenarios with varying difficulty levels, from simple rule applications to complex reasoning chains, in order to offer a precise evaluation of model performance on rule understanding and multi-step execution. Utilizing \model, we test various LLMs and identify notable shortcomings in their rule-based logical reasoning abilities.

\end{abstract}

\begin{figure}
    \centering
    \includegraphics[width=0.8\linewidth]{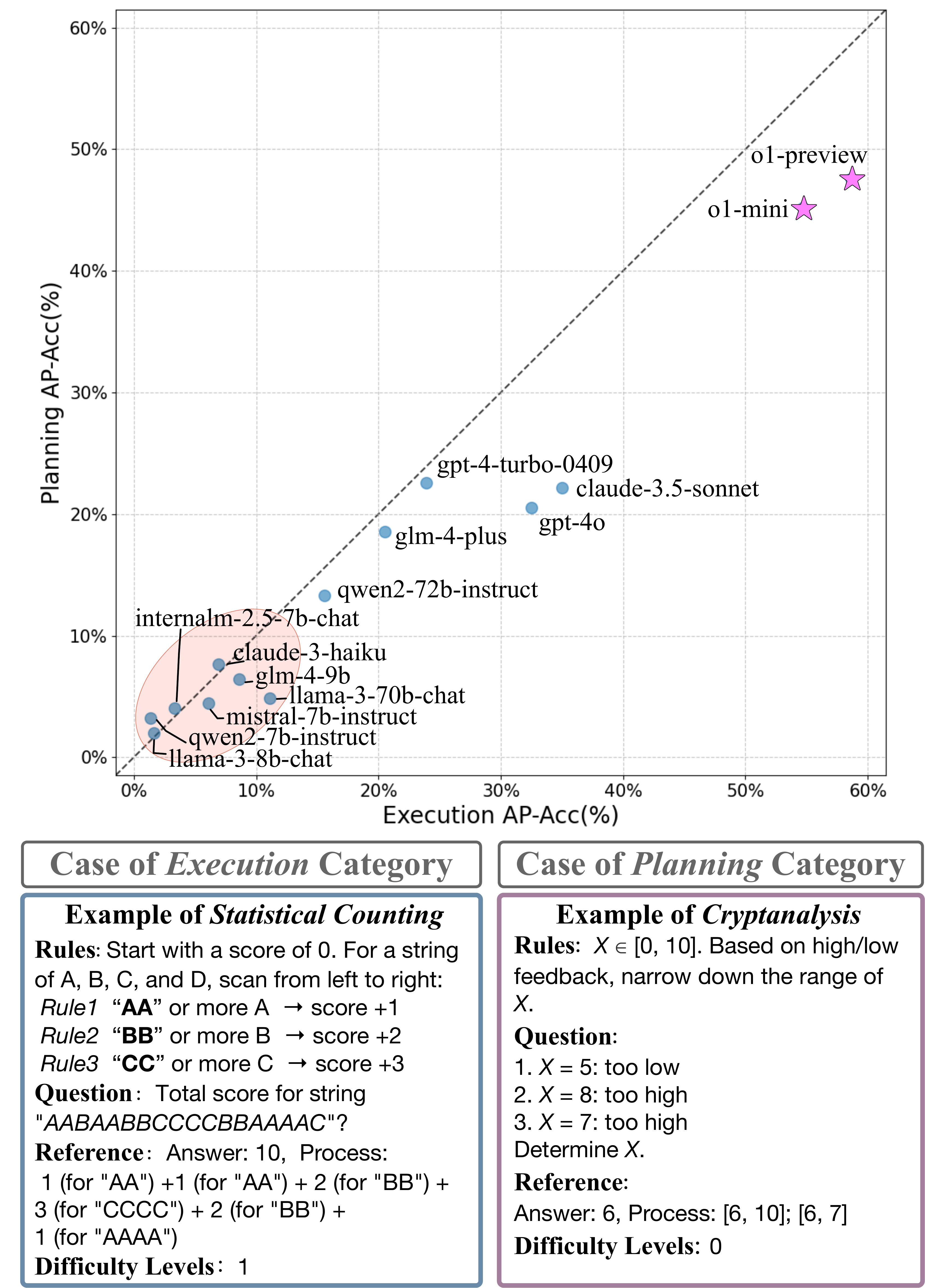}
    \caption{Evaluation results and demonstrations of \model. (Bottom) Case study on two examples from execution and planning category respectively. (Top) Performance of various models across execution and planning categories. The performance is arithmetic mean of \model's Chinese and English version. Most models struggle on \model getting less than 12\% scores in both categories. Two top-performing models highlighted with pink stars stand out.}
    \label{fig:intro}
\end{figure}

\section{Introduction}
    Large Language Models (LLMs) have shown notable abilities in a wide range of tasks and even complex problem-solving \cite{brown2020language, zeng2022glm, chowdhery2023palm, glm2024chatglm}. 
Their ability to reason, encompassing the understanding of intricate scenarios, strategic planning, and multi-step execution, is crucial for developing advanced AI agents and decision-making systems \cite{liu2023agentbench, sumers2023cognitive, cheng2024autodetect}.
These capabilities allow LLMs to understand complex user instructions, make logical decisions, and execute tasks accurately.

As alignment becomes integral to the application of LLMs \cite{chatgpt, Claude, ouyang2022training, cheng2023black}
, the primary goal is to align with human intentions and accurately execute their instructions.
Simultaneously, these models must possess strong reasoning abilities to handle complicated scenarios.
However, evaluating LLMs as effective rule-based executors and planners remains underexplored.
Traditional benchmarks usually focus solely on instruction-following or logical reasoning, neglecting the combination of both. Thus, they fail to comprehensively assess the model's reasoning capabilities after elaborate alignment.

In this paper, we introduce \textbf{\model}, a novel benchmark crafted to evaluate the comprehensive rule understanding, planning, and execution capabilities of LLMs.
\model offers a set of carefully designed rule-based reasoning games.
Each game contains a series of rules the model must follow to find the solution involving single or multiple steps. 
During the data construction process, we ensure that all the problems remain unavailable on the Internet to prevent data leakage.
\model covers two main scenarios: execution and planning, each divided into several sub-categories. 
Execution problems include tasks related to string data manipulation, where models handle string data transformations, as well as arithmetic operations and manipulations, focusing on mathematical computations and sequential execution. 
Planning games encompass math puzzles, which require solving complex mathematical problems, and pure logic puzzles, involving abstract reasoning without numerical computation. 
Through these varied games, \model aims to comprehensively evaluate the rule-based reasoning capabilities in LLMs.\footnote{We have released the dev set and the whole set of the \model, and created a leaderboard on https://github.com/Hypatiaalegra/LogicGame-Data. You can refer to https://www.codabench.org/competitions/4140/ for a fair and fast evaluation of the \model.}


In \model, our goal is to evaluate how well LLMs can reason according to given rules, so we ensure that no additional knowledge is required.
The final answer and the process in these scenarios rely solely on the given rules, fostering a pure assessment of the models' rule-based reasoning capabilities. 
Moreover, the evaluation process in \model involves not only the final answers but also the intermediate steps taken by the models, in order to offer a holistic view of the models' performance. 
Furthermore, process evaluation enables us to determine whether the model faithfully reasons based on established rules rather than merely guessing answers.
In addition, these intermediate steps are deterministic and can be automatically verified. 
To thoroughly assess the rule comprehension and multi-step execution capabilities of various LLMs, \model presents problems with multiple difficulty levels. 
We determine the complexity of each problem by the number of reasoning steps involved.
Simple games may require only single-step reasoning, while more challenging game scenarios require multiple reasoning steps, reflecting the need for deeper understanding and more sophisticated reasoning.

By leveraging \model, we have conducted extensive experiments across a wide range of LLMs, including api-based models like GPT and GLM families, as well as open-source models such as Qwen and Llama families.
Our findings indicate that while LLMs showcase good performance in a variety of tasks, they still exhibit notable shortcomings in rule-based logical reasoning. Even the best-performing LLMs struggle with complex reasoning tasks with about 20\% overall accuracy and less than 10\% on level 3 tasks. 
Additionally, while few-shot demonstrations can help with execution tasks, they may damage the performance of planning tasks.

Our contributions can be summarized as follows:
\begin{itemize}
    \item We introduce \model, a novel benchmark for rule-based reasoning, including execution and planning tasks, with varying difficulty levels.
    \item We design an automated assessment process for \model, which not only checks the final answers but also analyzes the solution process to comprehensively evaluate LLMs' reasoning abilities.
    \item We conduct extensive experiments on \model across a wide range of LLMs, effectively exposing their deficiencies in rule-based reasoning with the best about 25\% overall accuracy.
\end{itemize}

\section{Related Work}
    



The capability to reason has long been a crucial aspect of language models. Research \cite{wei2022emergent} has demonstrated that as the size of models increases, their ability to reason emerges, making it a fundamental attribute of LLMs. 
To elicit this reasoning ability, techniques like chain-of-thought prompting \cite{wei2022chain} and specialized training \cite{mukherjee2023orca} have become widely adopted.
Multi-step reasoning, in particular, is essential for complex decision-making and planning tasks, such as those undertaken by LLM agents \cite{liu2023agentbench}. 

Numerous benchmarks have been established over time to rigorously evaluate the reasoning capabilities of neural network models.
Early research has concentrated on logical reasoning \cite{bowman2013can, clark2020transformers, yu2020reclor}. These studies cover various forms of logic, including inductive, deductive, and abductive reasoning, and aim to assess whether models can infer answers based on given conditions.
Mathematical reasoning represents another critical area \cite{hendrycks2021measuring, mishra2022lila}. Benchmarks in this domain range in difficulty from grade school problems \cite{cobbe2021training} to Olympiad-level challenges \cite{huang2024olympicarena} and encompass a variety of formats, from word problems to theory proving \cite{li2020isarstep, lample2019deep}. These problems often demand not just reasoning but also robust calculation abilities.
Knowledge-based reasoning, particularly commonsense reasoning \cite{mishra2022numglue, onoe2021creak}, is another pivotal focus. These benchmarks are designed to determine whether models possess commonsense knowledge and can leverage it to reason effectively. Advancing further, theory-of-mind reasoning \cite{he2023hi} examines whether models can understand and incorporate complex layers of human cognition, such as thoughts and beliefs.

LLMs have undergone extensive alignment, with a key focus on following human instructions.
However, reasoning with the capability of instruction-following remains underexplored. Thus, we propose \model, a benchmark designed to assess rule-based reasoning, which is a natural integration of logical reasoning with instruction-following capabilities. We compare \model with each benchmark in Table~\ref{tab:comparison}.

\begin{table}[t]
    \centering
    \resizebox{\linewidth}{!}{ 
    \begin{tblr}{
        hline{1, Z} = {1pt, solid},
        hline{3, Y},
        hline{2} = {2-4}{solid},
        cell{1}{2} = {c = 3}{c},
        cell{1}{1} = {r = 2}{m},
        cell{1}{5-Z} = {r = 2}{m},
        cell{2}{2-4} = {m},
        cell{1-Z}{1-Z} = {c},
    }
        \textbf{Benchmark} & \textbf{Evaluation} & & &
         \textbf{Exemplars} & \textbf{Difficulty levels} &
        \textbf{Data Source} \\
        & \textbf{Process} & \textbf{Verifiable} & \textbf{Determinism} & & & & \\
        {\textbf{CLUTRR}~\cite{sinha2019clutrr}} & \xmark & \xmark & \xmark & \xmark & \xmark & semi-synthetic\\
        {\textbf{GSM8K}~\cite{cobbe2021training}} & \cmark & \cmark & \xmark & \xmark & \xmark & human-annotated\\
        {\textbf{PRONTOQA}~\cite{saparov2022language}} &\cmark & \cmark & \xmark & \xmark & \xmark & synthetic \\
        {\textbf{FOLIO}~\cite{han2022folio}} & \xmark & \xmark & \xmark & \cmark & \cmark & human-annotated \\
        {\textbf{BIG-Bench Hard}~\cite{suzgun2022challenging}} & \xmark & \xmark & \xmark & \cmark & \xmark & human-annotated \\
        {\textbf{STRATEGYQA}~\cite{geva2021did}} & \xmark & \xmark & \xmark & \xmark & \xmark & human-annotated \\
        \textbf{Ours} & \cmark & \cmark & \cmark & \cmark & \cmark & human-annotated \\
    \end{tblr}
    }
    \vspace{0.3cm} 
    \caption{Compare \model with other logical reasoning benchmarks. BIG-Bench Hard comparison limited to algorithmic part for relevance. Verifiable: Each step in process verifiable or not. Determinism: Each step in process determined or not. Verifiability and determinism guarantee automated evaluation of process. Exemplars: Examples provided or not.}
    \label{tab:comparison}
\end{table}

\section{\model}
    \subsection{Data Construction}
This section outlines our systematic approach to dataset construction, which comprises four key phases:
\begin{enumerate}
    \item Design rule-based problems inspired by real-world scenarios (\S\ref{para: problem design}).
    \item Develop output constraints to standardize evaluation formats (\S\ref{para: output constraint}).
    \item Implement a spectrum of difficulty levels and incorporating exemplars (\S\ref{para: difficulty}).
    \item Create bilingual versions through meticulous translation of problems and instructions (\S\ref{para: Bilingual}).
\end{enumerate}

\subsubsection{Problem Collection and Design} \label{para: problem design}


\paragraph{Collection and extraction of real-world scenarios.}
The integration of rule-following and reasoning is a critical aspect of many real-world tasks, yet existing benchmarks often fail to adequately capture this well. To address this gap, we developed a novel problem set through extensive research and crowdsourcing. We found that these tasks are similar to some game mechanics, since real-world tasks often share features with games, such as having specific rules to follow, requiring decision-making. This insight led us to adopt a gamification approach, allowing for a nuanced evaluation of models' rule-following reasoning capabilities.

\begin{figure}
    \centering
    \includegraphics[width=\linewidth]{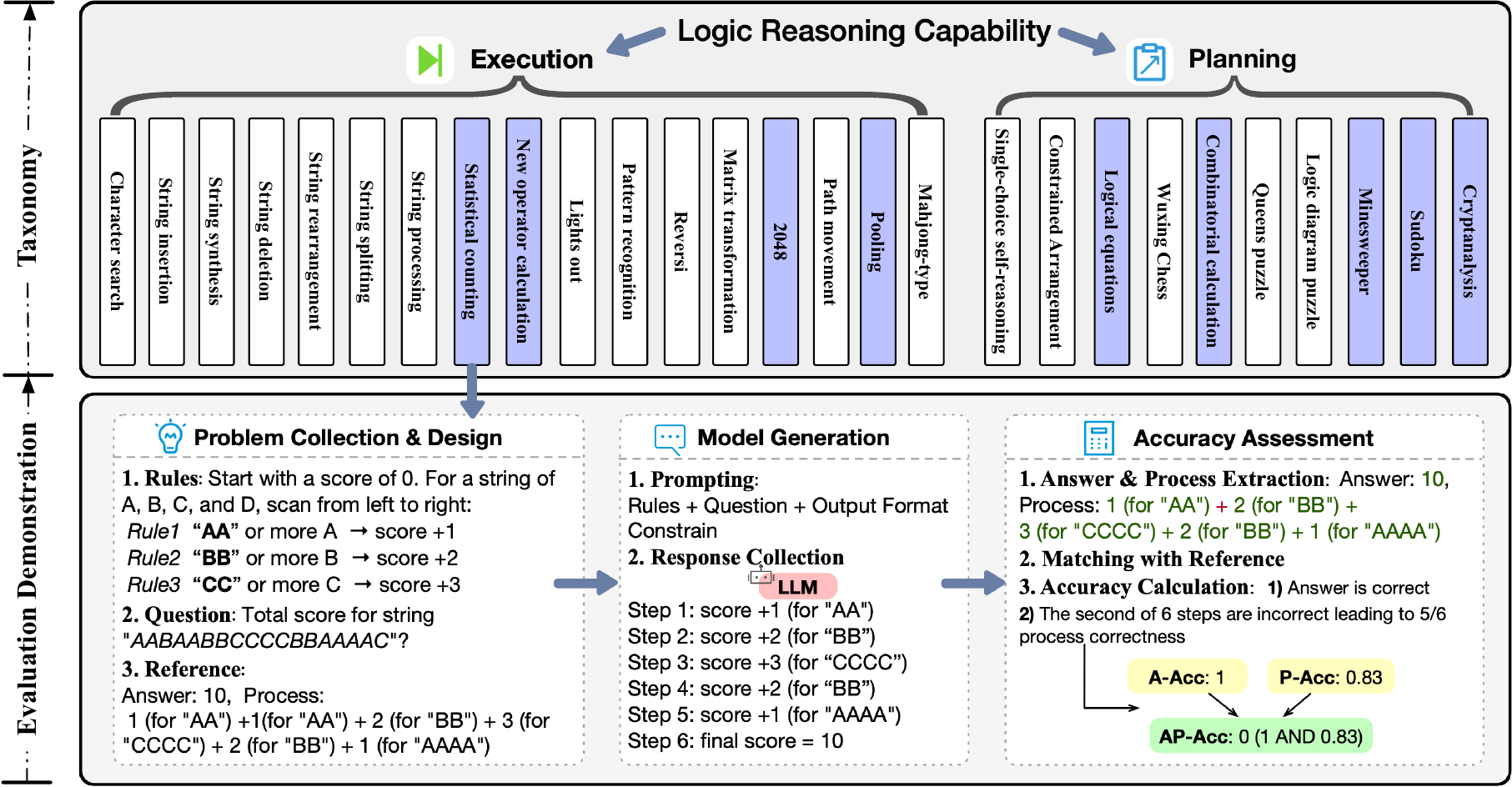}
    \caption{Illustration of taxonomy and evaluation protocol in \model. Taxonomy illustration highlights categories involving mathematics in purple. Json format constrain in evaluation is ommitted due to space limitations and can be referred to Appendix~\ref{sec:json prompt}.}
    \label{fig:main_process}
\end{figure}

The top part of Figure~\ref{fig:main_process} shows our categories. The dataset is structured into two primary domains: Execution and Planning.

\paragraph{Execution domain.}
In the context of our benchmark, execution refers to a reasoning process characterized by deterministic, single-step inferences. Here, models must apply well-defined rules to manipulate strings or states, with each step yielding a predictable outcome based on the current state and the applied rule. These tasks often require models to execute the correct action from given information, simulating real-world scenarios where explicit instructions are given.


\paragraph{Planning domain.}
The planning domain in our benchmark represents a higher order of cognitive complexity, involving long-term strategic thinking and multi-step decision making within rule-governed environments. Planning problems challenge models to analyze potential future states, formulate strategies, and determine a sequences of actions to reach a solution. Importantly, our focus in this domain is on identifying a correct solution path rather than optimizing for efficiency, mirroring many real-world scenarios where finding any valid solution is the primary goal.



\paragraph{Rule-based problem design and quality control.}
Following the establishment of our categories, a team of expert human annotators developed problems for each category with a focus on novelty and challenging out-of-domain reasoning, which makes it harder to overfit.
To mitigate potential semantic ambiguities associated with natural language reasoning \cite{fedorenko2024language}, we minimized reliance on natural language constructs. Our problems are designed such that the reasoning process does not necessitate natural language inference, allowing for a more direct evaluation of reasoning abilities.

In the execution domain, we ensured that every step is deterministic and verifiable, facilitating precise evaluation and preventing models from resorting to guesswork. For the inherently less deterministic planning problems, we introduced intermediate checkpoints or state variables where appropriate, allowing for a more granular assessment of the problem-solving process. Detailed specifications of these evaluation methods will be provided in Section~\ref{sec: evaluation_method}.


\subsubsection{Output Constraint Design} \label{para: output constraint}
To facilitate precise evaluation and streamline the matching process, we mandated a structured JSON output format for model responses. 
Our evaluation criteria are tailored to the complexity of each problem. For single-step problems categorized as Level 0, models are only required to output the final answer, and evaluation is based solely on the correctness of this answer. However, for problems involving multiple steps or more complex reasoning, which include Levels 1, 2, 3, and certain Level 0 problems, we evaluate both the answer and the process.

In both cases, the output JSON structure includes 'answer', which is a list of strings representing the final solution(s), and for second cases the output also includes 'process', a list of strings detailing each step of the problem-solving process. The details of JSON constraints can be found in Appendix ~\ref{sec:json prompt}

\subsubsection{Difficulty Levels and Exemplars} \label{para: difficulty}
To comprehensively assess models' reasoning capabilities, we have structured our benchmark with four distinct difficulty levels (0, 1, 2, and 3) for each task. The difficulty gradient is determined by two key factors: the complexity of the rules involved and the number of reasoning steps required to arrive at the solution. Each successive level systematically introduces additional rules and reasoning steps. In general, our problems are difficult for models, and some are also challenging for humans.

Furthermore, to evaluate models' capacity for rule acquisition and application, we have developed two distinct exemplars for each question. These exemplars consist of a given question, the correct answer, a step-by-step solution process, and detailed explanations. By providing these examples, we aim to test not only the models' baseline performance but also their ability to learn from demonstrations and apply newly acquired rules to similar problems.

\subsubsection{Building Bilingual Benchmark} \label{para: Bilingual}
Initially, our questions were designed in Chinese. However, we recognized that this could potentially bias the benchmark against LLMs primarily trained on English data. To ensure fairness and broader applicability of our benchmark, we developed a comprehensive bilingual benchmark containing both \textit{zh}(Chinese) and \textit{en}(English) version.


\subsection{Evaluation Protocol}
\label{sec: evaluation_method}

Each model is prompted with a set of rules specific to the given problem, along with a corresponding question and a JSON format constraint for the output, encompassing both the answer and the process, as illustrated in Figure~\ref{fig:main_process}. For few-shot trials, example(s) are inserted between the rules and the question to assess the model's in-context learning capabilities. The model responses are then collected and subjected to automated evaluation. As previously mentioned, the evaluation protocol is designed to assess not only the correctness of the answer but also the correctness of the process that led to the answer. Scoring for each problem's answer is determined by comparing the model's response to the reference answer. Similarly, scoring for each problem's process, as defined by the JSON format constraint, is achieved by assessing the degree of alignment between the model's process and the reference process. Specifically, the \model defines three metrics related to each problem for scoring:
\begin{enumerate}
    \item \textbf{Answer Accuracy(A-Acc)}: This metric evaluates the correctness of the answers for all given questions, providing a binary assessment (0/1) for each answer to indicate whether it is correct or not.
    \item \textbf{Process Accuracy(P-Acc):} This metric assesses the correctness of the process, measuring the percentage match based on character-level similarity between the provided process and the expected process. In rare cases where no process is provided in level 0 questions as single-step reasoning, process accuracy are considered equally with answer accuracy for scoring.
    \item \textbf{Answer Process Accuracy(AP-Acc):} This composite metric evaluates the overall accuracy of both the answer and the process. Its calculation involves an aggregate score derived by combining answer accuracy and process accuracy using a logical AND operation.
\end{enumerate}

\section{Experiments}
    \subsection{Experimental Setup}
We evaluate 14 popular LLMs. Closed-source models included versions from Claude \cite{Claude} and GPT \cite{openai2023gpt} series. Open-source models encompassed LLaMA 3 \cite{touvron2023llama}, Qwen \cite{bai2023qwen}, GLM \cite{glm2024chatglm}, Mistral \cite{jiang2023mistral}, and InternLM \cite{2023internlm} variants.
In the inference stage, we set temperature to 0, ensuring deterministic outputs. The maximum token number is set to 2048. Other parameters are set as their default values.

\begin{figure}
    \centering
    \begin{subfigure}{\textwidth}
        \centering
        \small
        \resizebox{\textwidth}{!}{
        \begin{tblr}{
            hline{1, Z} = {1pt, solid},
            hline{2, 3},
            cell{1}{1} = {r = 2}{m},
            cell{1}{2} = {c = 5}{c},
            cell{1}{7} = {c = 5}{c},
            cell{1}{12} = {r = 2}{m},
            vline{2, 6, 7, 11, 12},
        }
            \textbf{Model} & \textbf{Execution} & & & & & \textbf{Planning} & & & & & \textbf{Overall} \\
             & \textbf{Level 0} & \textbf{Level 1} & \textbf{Level 2} & \textbf{Level 3} & \textbf{Avg.} & \textbf{Level 0} & \textbf{Level 1} & \textbf{Level 2} & \textbf{Level 3} & \textbf{Avg.} & \\
            \textbf{o1-preview} & \textbf{82.22} & \textbf{66.67} & \textbf{48.89} & \textbf{42.22} & \textbf{60.00} & \textbf{64.52} & \textbf{58.06} & \textbf{35.48} & \textbf{32.26} & \textbf{47.58} & \textbf{54.93} \\
            \textbf{o1-mini} & 77.78 & 57.78 & \textbf{48.89} & \textbf{42.22} & 56.67 & 74.19 & \textbf{58.06} & \textbf{35.48} & 12.90 & 45.16 & 51.97 \\
            \textbf{claude-3-5-sonnet} & 68.89 & 40.00 & 22.22 & 15.56 & 36.67 & 48.39 & 25.81 & 6.45 & 3.23 & 20.97 & 30.26 \\
            \textbf{gpt-4o} & 62.22 & 31.11 & 13.33 & 13.33 & 30.00 & 51.61 & 22.58 & 9.68 & 6.45 & 22.58 & 26.97 \\
            \textbf{gpt-4-turbo-0409} & 60.00 & 15.56 & 15.56 & 11.11 & 25.56 & 67.74 & 22.58 & 9.68 & 3.23 & 25.81 & 25.66 \\
            \textbf{glm-4-plus} & 42.22 & 26.67 & 17.78 & 6.67 & 23.33 & 48.39 & 19.35 & 6.45 & 3.23 & 19.36 & 21.71 \\
            \textbf{qwen2-72b} & 53.33 & 20.00 & 15.56 & 2.22 & 22.78 & 45.16 & 16.13 & 3.23 & 3.23 & 16.94 & 20.39 \\
            \textbf{llama-3-70b} & 42.22 & 11.11 & 6.67 & 2.22 & 15.56 & 25.81 & 6.45 & 0.00 & 0.00 & 8.07 & 12.50 \\
            \textbf{claude-3-haiku} & 31.11 & 4.44 & 2.22 & 0.00 & 9.44 & 32.26 & 6.45 & 0.00 & 0.00 & 9.68 & 9.54 \\
            \textbf{glm-4-9b} & 24.44 & 8.89 & 2.22 & 0.00 & 8.89 & 19.35 & 3.23 & 0.00 & 0.00 & 5.65 & 7.57 \\
            \textbf{internlm2-5-7b} & 13.33 & 4.44 & 0.00 & 0.00 & 4.44 & 16.13 & 3.23 & 0.00 & 0.00 & 4.84 & 4.61 \\
            \textbf{llama-3-8b} & 11.11 & 2.22 & 0.00 & 0.00 & 3.33 & 9.68 & 6.45 & 0.00 & 0.00 & 4.03 & 3.62 \\
            \textbf{mistral-7b} & 4.44 & 0.00 & 0.00 & 0.00 & 1.11 & 19.35 & 3.23 & 0.00 & 0.00 & 5.65 & 2.96 \\
            \textbf{qwen2-7b} & 4.44 & 0.00 & 0.00 & 0.00 & 1.11 & 16.13 & 3.23 & 0.00 & 0.00 & 4.84 & 2.63 \\
        \end{tblr}
        }
        \caption{Performance of 14 models on \model of \textit{zh} version. The highest performance is \textbf{bold}.}
        \label{tab:main_result_zh}
    \end{subfigure}
    
    \vspace{1cm}
    
    \begin{subfigure}{\textwidth}
        \centering
        \small
        \resizebox{\textwidth}{!}{
        \begin{tblr}{
            hline{1, Z} = {1pt, solid},
            hline{2, 3},
            cell{1}{1} = {r = 2}{m},
            cell{1}{2} = {c = 5}{c},
            cell{1}{7} = {c = 5}{c},
            cell{1}{12} = {r = 2}{m},
            vline{2, 6, 7, 11, 12},
        }
            \textbf{Model} & \textbf{Execution} & & & & & \textbf{Planning} & & & & & \textbf{Overall} \\
             & \textbf{Level 0} & \textbf{Level 1} & \textbf{Level 2} & \textbf{Level 3} & \textbf{Avg.} & \textbf{Level 0} & \textbf{Level 1} & \textbf{Level 2} & \textbf{Level 3} & \textbf{Avg.} & \\
            \textbf{o1-preview} & 71.11 & \textbf{57.78} & \textbf{48.89} & \textbf{51.11} & \textbf{57.22} & \textbf{70.97} & \textbf{54.84} & \textbf{41.94} & \textbf{22.58} & \textbf{47.58} & \textbf{53.29} \\
            \textbf{o1-mini} & \textbf{73.33} & 46.67 & 44.44 & 46.67 & 52.78 & \textbf{70.97} & 48.39 & 38.71 & \textbf{22.58} & 45.16 & 49.67 \\
            \textbf{claude-3-5-sonnet} & 46.67 & 40.00 & 33.33 & 13.33 & 33.33 & 64.52 & 16.13 & 6.45 & 6.45 & 23.39 & 29.28 \\
            \textbf{gpt-4o} & 57.78 & 37.78 & 31.11 & 13.33 & 35.00 & 48.39 & 19.35 & 3.23 & 3.23 & 18.55 & 28.29 \\
            \textbf{gpt-4-turbo-0409} & 42.22 & 20.00 & 17.78 & 8.89 & 22.22 & 51.61 & 12.90 & 9.68 & 3.23 & 19.36 & 21.05 \\
            \textbf{glm-4-plus} & 31.11 & 15.56 & 17.78 & 6.67 & 17.78 & 48.39 & 9.68 & 9.68 & 3.23 & 17.75 & 17.76 \\
            \textbf{qwen2-72b} & 28.89 & 4.44 & 0.00 & 0.00 & 8.33 & 22.58 & 6.45 & 6.45 & 3.23 & 9.68 & 8.88 \\
            \textbf{glm-4-9b} & 22.22 & 6.67 & 2.22 & 2.22 & 8.33 & 22.58 & 6.45 & 0.00 & 0.00 & 7.26 & 7.89 \\
            \textbf{internlm2-5-7b} & 22.22 & 4.44 & 4.44 & 0.00 & 7.78 & 12.90 & 3.23 & 0.00 & 0.00 & 4.03 & 6.25 \\
            \textbf{claude-3-haiku} & 8.89 & 8.89 & 0.00 & 0.00 & 4.45 & 22.58 & 0.00 & 0.00 & 0.00 & 5.65 & 4.93 \\
            \textbf{llama-3-70b} & 8.89 & 8.89 & 8.89 & 0.00 & 6.67 & 3.23 & 3.23 & 0.00 & 0.00 & 1.62 & 4.61 \\
            \textbf{mistral-7b} & 22.22 & 0.00 & 0.00 & 0.00 & 5.56 & 9.68 & 0.00 & 0.00 & 0.00 & 2.42 & 4.28 \\
            \textbf{qwen2-7b} & 4.44 & 2.22 & 0.00 & 0.00 & 1.67 & 6.45 & 0.00 & 0.00 & 0.00 & 1.61 & 1.64 \\
            \textbf{llama-3-8b} & 0.00 & 0.00 & 0.00 & 0.00 & 0.00 & 0.00 & 0.00 & 0.00 & 0.00 & 0.00 & 0.00 \\
        \end{tblr}
        }
        \caption{Performance of 14 models on \model of \textit{en} version. The highest performance is \textbf{bold}.}
        \label{tab:main_result_en}
    \end{subfigure}
    
    \caption{Performance comparison of 14 models on \model measured by AP-Acc for both Chinese (zh) and English (en) versions.}
    \label{fig:combined_results}
\end{figure}

\subsection{Main Results and Analysis}

Table~\ref{fig:combined_results} presents the performance of 14 LLMs on \model across our Chinese and English bilingual dataset, measured by AP-Acc. The o1-preview model leads with 54.93\% and 53.29\% overall accuracy for Chinese and English respectively, closely followed by o1-mini. These results underscore the persistent challenge of reasoning for LLMs, as even top performers barely exceed 50\% accuracy. The substantial performance gap, ranging from over 50\% to below 5\%, not only highlights the varying capabilities of current LLMs in complex reasoning tasks, but also emphasizes that despite recent advancements, logical reasoning remains a significant hurdle for most language models.

The performance drop from Level 0 to Level 3 is not uniform across models. Some models (e.g. o1-mini) show a more gradual decline, while others drop off sharply after Level 1. This may suggest that some models have better consistency across task complexities.

Interestingly, the performance of models in Execution and Planning tasks varies. Some top-performing models, such as o1-preview,  demonstrates superior performance in Execution compared to Planning. Conversely, llama-3-70b-chat excels in Planning over Execution. Even within model families, the relative performance differences are evident. In the GPT family, gpt-4-turbo-0409 outperforms gpt-4o in Planning tasks, while gpt-4o shows better performance in Execution tasks.

During the evaluation, it was observed that some models occasionally failed to adhere to the requirement of producing JSON format output. Detailed results and analysis are provided in Appendix~\ref{sec:data_overview}. Despite this, the overall error rates are low across most models, resulting in minimal differences in rankings among those with similar performance.

\subsection{Few-shot Results}
\begin{figure}
    \centering
    \includegraphics[width=\linewidth]{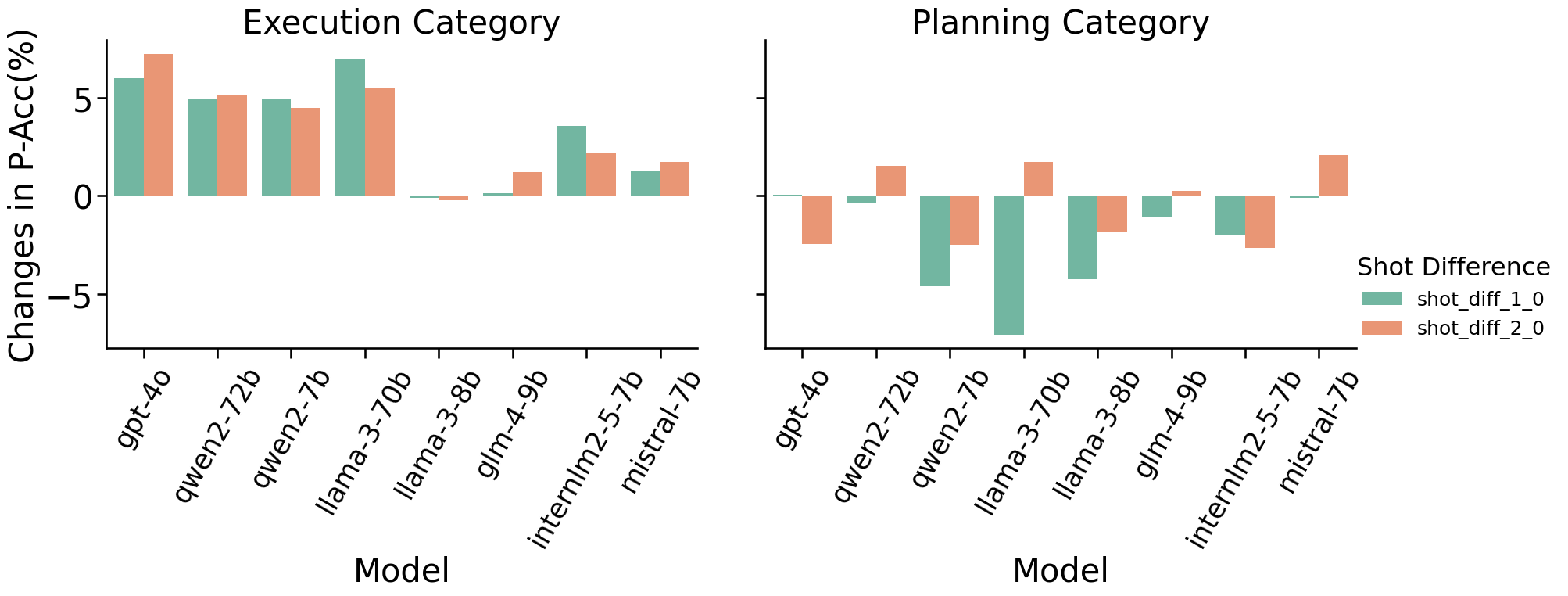}
    \caption{Few-shot differences on execution and planning category of \model's \textit{zh} version. "shot\_diff\_1\_0" represents the difference in the P-Acc score between the 1-shot and 0-shot settings, calculated as the result of 1-shot minus the result of 0-shot, "shot\_diff\_2\_0" representing the P-Acc score between the 2-shot and 0-shot settings similarly.}
    \label{fig:category_diff_fewshot}
\end{figure}

\begin{figure}
    \centering
    \includegraphics[width=\linewidth]{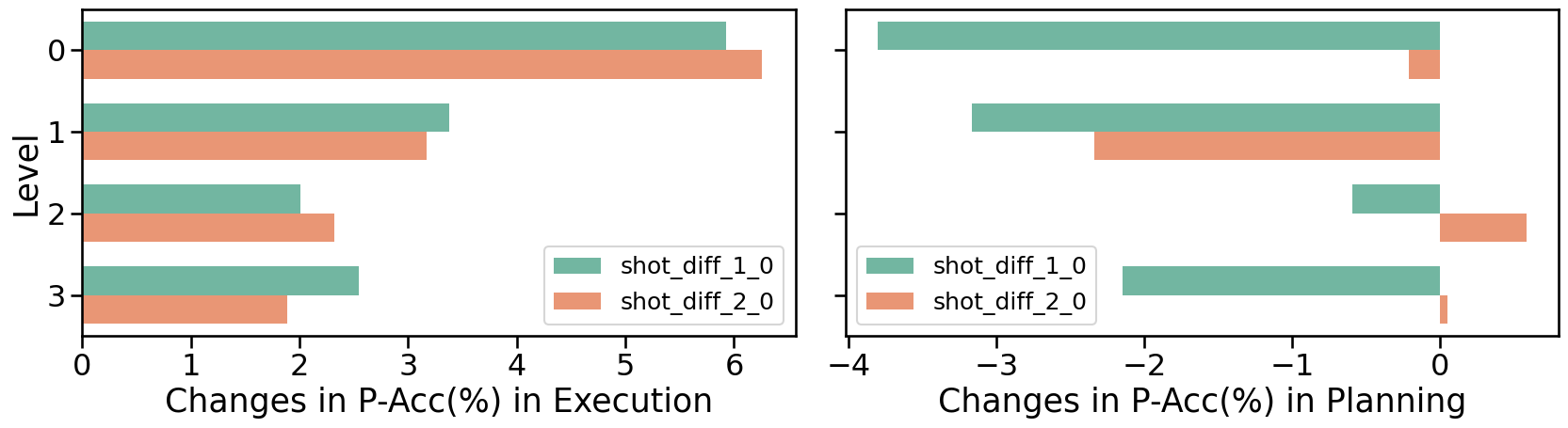}
    \caption{Few-shot differences on difficulty levels of \model's \textit{zh} version with shot difference settings similar with Figure~\ref{fig:category_diff_fewshot}.}
    \label{fig:level_diff_fewshot}
\end{figure}

We conducted experiments to analyze the change of model's performance on 0-shot, 1-shot and 2-shot settings. And models of gpt-4o, qwen2-72b-instruct and qwen2-7b-instruct, llama-3-70b-chat and llama-3-8b-chat, glm-4-9b, mistral-7b-instruct and internlm-2.5-7b-chat are chosen for this trial. 
The analysis demonstrated in Figure~\ref{fig:category_diff_fewshot} and Figure~\ref{fig:level_diff_fewshot} reveals mixed results of \model's \textit{zh} version in the "Planning" and "Execution" categories, difficulty levels across the various shot settings. Appendix~\ref{sec:en_few_shot} shows the results of \model's en version.

\begin{figure}
    \centering
    \includegraphics[width=0.8\linewidth]{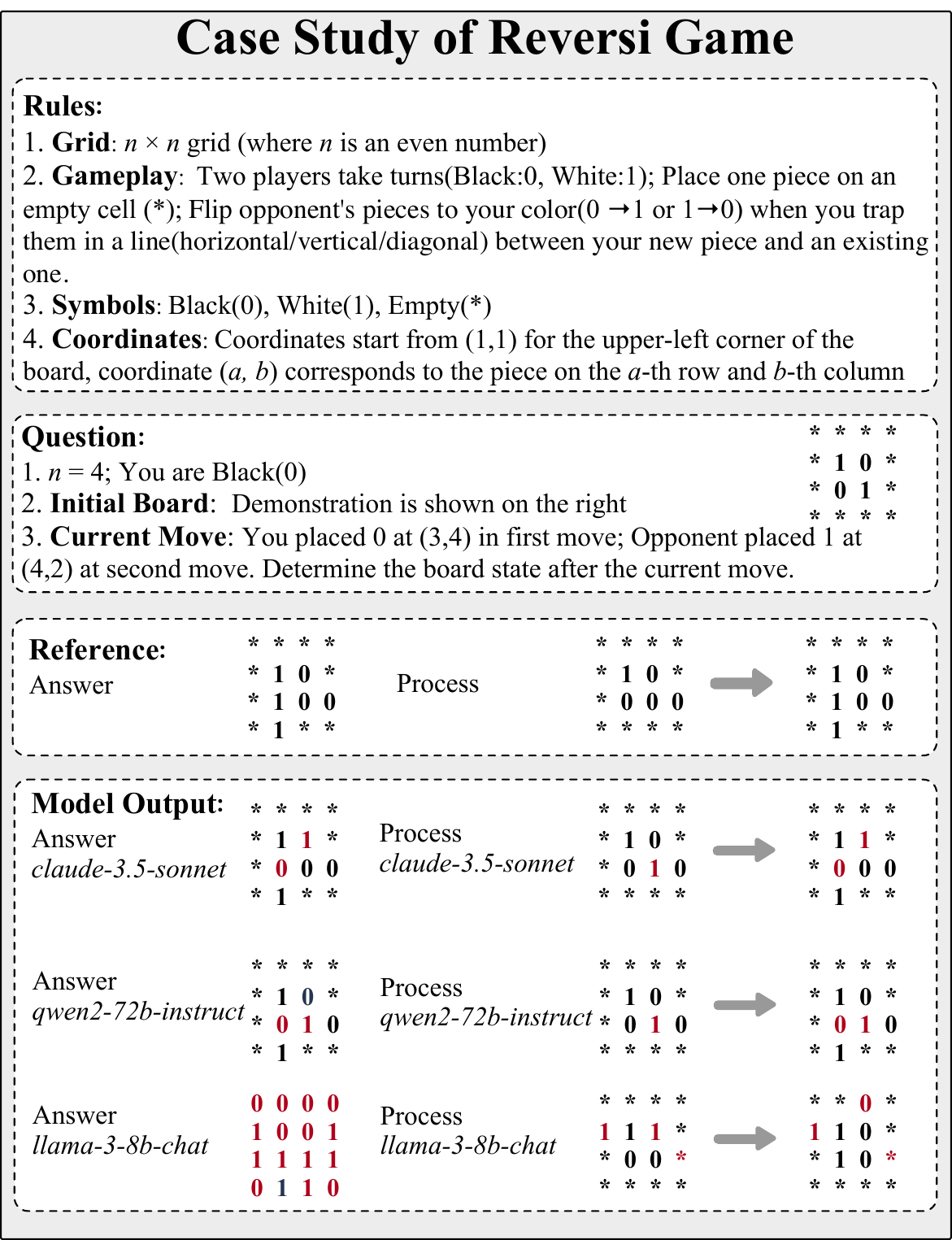}
    \caption{An example of a Reversi game with model outputs, including the answer and process, is shown. The board states for initial, reference, and model outputs are visualized with errors highlighted in red. JSON constraints are omitted due to space, referenced in Figure~\ref{fig:json-prompt}.}
    \label{fig:case_study}
\end{figure}

In the "Execution" category, models demonstrate notable improvements in accuracy with increased shot contexts demonstrated in Figure~\ref{fig:category_diff_fewshot}. Specifically, \textbf{stronger models }(as indicated in Table~\ref{tab:main_result_zh} and Table~\ref{tab:main_result_en}) like gpt-4o, qwen2-72b-instruct \textbf{shows a greater increase in the AP-Acc score} when transitioning from \textbf{0-shot to 1-shot and 2-shot} settings than weaker ones, indicating enhanced execution accuracy with more contextual information. However, the effects of 1-shot and 2-shot settings vary across models
Performance variations by difficulty levels, as shown in Figure~\ref{fig:level_diff_fewshot}, indicate that \textbf{models benefit most from 1-shot and 2-shot contexts at Level 0}. And in general, \textbf{the influence of shot contexts diminishes as the difficulty level increases}. This consistency suggests that simpler tasks (Level 0) allow models to leverage additional context more effectively, enhancing their execution capabilities across the board.

Conversely, the "Planning" category presents more heterogeneous results. \textbf{Models often show declines in performance when moving from 0-shot to 1-shot or 2-shot} settings demonstrated in Figure~\ref{fig:category_diff_fewshot}. These results suggest that while additional context can enhance performance for some models, it can introduce noise for others, potentially obfuscating key elements necessary for planning tasks. Overall, these observations highlight 
that \textbf{the added context's efficacy is highly contingent on the task and model characteristics}. Furthermore, Figure~\ref{fig:level_diff_fewshot} illustrates that the \textbf{negative impact of 1-shot contexts is most pronounced at Level 0. As the difficulty level increases, the influence of 1-shot contexts diminishes}, leading to smaller performance fluctuations, while 2-shot contexts are more unstable with no pattern found.


\section{Discussion}

\paragraph{Case study on Revesi game.}  From Appendix~\ref{sec:fail_heatmap},  it is evident that all models perform poorly in the Reversi game. Consequently, we conducted a case study on this particular game scenario. The responses from various models tasked with determining the outcome of a Reversi game are analyzed as shown in Figure~\ref{fig:case_study}. Despite all models except llama-3-8b-chat adhering to the instruction format and correctly interpreting the initial setup, all models failed to provide the correct answer, demonstrating various types of inaccuracies. The key reasons for failure include:
\begin{enumerate}
    \item \textbf{Mismanagement of Details}: For instance, claude-3.5-sonnet misplaced markers or incorrectly transformed pieces, showing that while the general rules were understood, the model failed to apply specific game rules correctly.
    \item \textbf{Inadequate Execution/Planning Understanding}: Models like qwen2-72b-instruct produced incorrect board states following what should have been straightforward captures, revealing a fundamental misunderstanding of the game's piece-flipping mechanisms as well as the initial conditions outlined in the problem.
    \item \textbf{Excessive Alterations}: The llama-3-8b-chat model drastically altered the board state in an unrealistic manner, adding rows and altering more positions than the rules allow, suggesting a misinterpretation of the core principles of the game, particularly with regards to matrix operations and the understanding and execution of piece-flipping mechanisms.
\end{enumerate}

\section{Conclusion}
    In this paper, we introduce \model, a novel benchmark designed to evaluate the rule-based reasoning capabilities of LLMs. \model encompasses multiple difficulty levels, focusing on assessing models' understanding of rules, execution based on these rules, and planning abilities.
Moreover, we have developed methods to evaluate both outcomes and reasoning processes, ensuring that models follow the given rules faithfully rather than merely guessing answers. Extensive experiments indicate that current large models still exhibit significant deficiencies in rule-based reasoning tasks. More effort needs to be devoted to further enhancing models' abilities to handle complex reasoning scenarios.

\clearpage

\bibliographystyle{abbrv}
\bibliography{ref_back}

\begin{thebibliography}{10}

\bibitem{Claude}
Anthropic.
\newblock Introducing claude, 2023.

\bibitem{bai2023qwen}
J.~Bai, S.~Bai, Y.~Chu, Z.~Cui, K.~Dang, X.~Deng, Y.~Fan, W.~Ge, Y.~Han, F.~Huang, et~al.
\newblock Qwen technical report.
\newblock {\em arXiv preprint arXiv:2309.16609}, 2023.

\bibitem{bowman2013can}
S.~R. Bowman.
\newblock Can recursive neural tensor networks learn logical reasoning?
\newblock {\em arXiv preprint arXiv:1312.6192}, 2013.

\bibitem{brown2020language}
T.~Brown, B.~Mann, N.~Ryder, M.~Subbiah, J.~D. Kaplan, P.~Dhariwal, A.~Neelakantan, P.~Shyam, G.~Sastry, A.~Askell, et~al.
\newblock Language models are few-shot learners.
\newblock {\em Advances in neural information processing systems}, 33:1877--1901, 2020.

\bibitem{cheng2023black}
J.~Cheng, X.~Liu, K.~Zheng, P.~Ke, H.~Wang, Y.~Dong, J.~Tang, and M.~Huang.
\newblock Black-box prompt optimization: Aligning large language models without model training.
\newblock {\em arXiv preprint arXiv:2311.04155}, 2023.

\bibitem{cheng2024autodetect}
J.~Cheng, Y.~Lu, X.~Gu, P.~Ke, X.~Liu, Y.~Dong, H.~Wang, J.~Tang, and M.~Huang.
\newblock Autodetect: Towards a unified framework for automated weakness detection in large language models.
\newblock {\em arXiv preprint arXiv:2406.16714}, 2024.

\bibitem{chowdhery2023palm}
A.~Chowdhery, S.~Narang, J.~Devlin, M.~Bosma, G.~Mishra, A.~Roberts, P.~Barham, H.~W. Chung, C.~Sutton, S.~Gehrmann, et~al.
\newblock Palm: Scaling language modeling with pathways.
\newblock {\em Journal of Machine Learning Research}, 24(240):1--113, 2023.

\bibitem{clark2020transformers}
P.~Clark, O.~Tafjord, and K.~Richardson.
\newblock Transformers as soft reasoners over language.
\newblock {\em arXiv preprint arXiv:2002.05867}, 2020.

\bibitem{cobbe2021training}
K.~Cobbe, V.~Kosaraju, M.~Bavarian, M.~Chen, H.~Jun, L.~Kaiser, M.~Plappert, J.~Tworek, J.~Hilton, R.~Nakano, et~al.
\newblock Training verifiers to solve math word problems.
\newblock {\em arXiv preprint arXiv:2110.14168}, 2021.

\bibitem{fedorenko2024language}
E.~Fedorenko, S.~T. Piantadosi, and E.~A. Gibson.
\newblock Language is primarily a tool for communication rather than thought.
\newblock {\em Nature}, 630(8017):575--586, 2024.

\bibitem{geva2021did}
M.~Geva, D.~Khashabi, E.~Segal, T.~Khot, D.~Roth, and J.~Berant.
\newblock Did aristotle use a laptop? a question answering benchmark with implicit reasoning strategies.
\newblock {\em Transactions of the Association for Computational Linguistics}, 9:346--361, 2021.

\bibitem{glm2024chatglm}
T.~GLM, :, A.~Zeng, B.~Xu, B.~Wang, C.~Zhang, D.~Yin, D.~Rojas, G.~Feng, H.~Zhao, H.~Lai, H.~Yu, H.~Wang, J.~Sun, J.~Zhang, J.~Cheng, J.~Gui, J.~Tang, J.~Zhang, J.~Li, L.~Zhao, L.~Wu, L.~Zhong, M.~Liu, M.~Huang, P.~Zhang, Q.~Zheng, R.~Lu, S.~Duan, S.~Zhang, S.~Cao, S.~Yang, W.~L. Tam, W.~Zhao, X.~Liu, X.~Xia, X.~Zhang, X.~Gu, X.~Lv, X.~Liu, X.~Liu, X.~Yang, X.~Song, X.~Zhang, Y.~An, Y.~Xu, Y.~Niu, Y.~Yang, Y.~Li, Y.~Bai, Y.~Dong, Z.~Qi, Z.~Wang, Z.~Yang, Z.~Du, Z.~Hou, and Z.~Wang.
\newblock Chatglm: A family of large language models from glm-130b to glm-4 all tools, 2024.

\bibitem{han2022folio}
S.~Han, H.~Schoelkopf, Y.~Zhao, Z.~Qi, M.~Riddell, L.~Benson, L.~Sun, E.~Zubova, Y.~Qiao, M.~Burtell, et~al.
\newblock Folio: Natural language reasoning with first-order logic.
\newblock {\em arXiv preprint arXiv:2209.00840}, 2022.

\bibitem{he2023hi}
Y.~He, Y.~Wu, Y.~Jia, R.~Mihalcea, Y.~Chen, and N.~Deng.
\newblock Hi-tom: A benchmark for evaluating higher-order theory of mind reasoning in large language models.
\newblock {\em arXiv preprint arXiv:2310.16755}, 2023.

\bibitem{hendrycks2021measuring}
D.~Hendrycks, C.~Burns, S.~Basart, A.~Zou, M.~Mazeika, D.~Song, and J.~Steinhardt.
\newblock Measuring massive multitask language understanding.
\newblock In {\em International Conference on Learning Representations}, 2021.

\bibitem{huang2024olympicarena}
Z.~Huang, Z.~Wang, S.~Xia, X.~Li, H.~Zou, R.~Xu, R.-Z. Fan, L.~Ye, E.~Chern, Y.~Ye, et~al.
\newblock Olympicarena: Benchmarking multi-discipline cognitive reasoning for superintelligent ai.
\newblock {\em arXiv preprint arXiv:2406.12753}, 2024.

\bibitem{jiang2023mistral}
A.~Q. Jiang, A.~Sablayrolles, A.~Mensch, C.~Bamford, D.~S. Chaplot, D.~d.~l. Casas, F.~Bressand, G.~Lengyel, G.~Lample, L.~Saulnier, et~al.
\newblock Mistral 7b.
\newblock {\em arXiv preprint arXiv:2310.06825}, 2023.

\bibitem{lample2019deep}
G.~Lample and F.~Charton.
\newblock Deep learning for symbolic mathematics.
\newblock {\em arXiv preprint arXiv:1912.01412}, 2019.

\bibitem{li2020isarstep}
W.~Li, L.~Yu, Y.~Wu, and L.~C. Paulson.
\newblock Isarstep: a benchmark for high-level mathematical reasoning.
\newblock {\em arXiv preprint arXiv:2006.09265}, 2020.

\bibitem{liu2023agentbench}
X.~Liu, H.~Yu, H.~Zhang, Y.~Xu, X.~Lei, H.~Lai, Y.~Gu, H.~Ding, K.~Men, K.~Yang, et~al.
\newblock Agentbench: Evaluating llms as agents.
\newblock {\em arXiv preprint arXiv:2308.03688}, 2023.

\bibitem{mishra2022lila}
S.~Mishra, M.~Finlayson, P.~Lu, L.~Tang, S.~Welleck, C.~Baral, T.~Rajpurohit, O.~Tafjord, A.~Sabharwal, P.~Clark, et~al.
\newblock Lila: A unified benchmark for mathematical reasoning.
\newblock {\em arXiv preprint arXiv:2210.17517}, 2022.

\bibitem{mishra2022numglue}
S.~Mishra, A.~Mitra, N.~Varshney, B.~Sachdeva, P.~Clark, C.~Baral, and A.~Kalyan.
\newblock Numglue: A suite of fundamental yet challenging mathematical reasoning tasks.
\newblock {\em arXiv preprint arXiv:2204.05660}, 2022.

\bibitem{mukherjee2023orca}
S.~Mukherjee, A.~Mitra, G.~Jawahar, S.~Agarwal, H.~Palangi, and A.~Awadallah.
\newblock Orca: Progressive learning from complex explanation traces of gpt-4.
\newblock {\em arXiv preprint arXiv:2306.02707}, 2023.

\bibitem{onoe2021creak}
Y.~Onoe, M.~J. Zhang, E.~Choi, and G.~Durrett.
\newblock Creak: A dataset for commonsense reasoning over entity knowledge.
\newblock {\em arXiv preprint arXiv:2109.01653}, 2021.

\bibitem{chatgpt}
OpenAI.
\newblock Introducing chatgpt, 2022.

\bibitem{openai2023gpt}
R.~OpenAI.
\newblock Gpt-4 technical report.
\newblock {\em arXiv}, pages 2303--08774, 2023.

\bibitem{ouyang2022training}
L.~Ouyang, J.~Wu, X.~Jiang, D.~Almeida, C.~Wainwright, P.~Mishkin, C.~Zhang, S.~Agarwal, K.~Slama, A.~Ray, et~al.
\newblock Training language models to follow instructions with human feedback.
\newblock {\em Advances in Neural Information Processing Systems}, 35:27730--27744, 2022.

\bibitem{saparov2022language}
A.~Saparov and H.~He.
\newblock Language models are greedy reasoners: A systematic formal analysis of chain-of-thought.
\newblock {\em arXiv preprint arXiv:2210.01240}, 2022.

\bibitem{sinha2019clutrr}
K.~Sinha, S.~Sodhani, J.~Dong, J.~Pineau, and W.~L. Hamilton.
\newblock Clutrr: A diagnostic benchmark for inductive reasoning from text.
\newblock {\em arXiv preprint arXiv:1908.06177}, 2019.

\bibitem{sumers2023cognitive}
T.~R. Sumers, S.~Yao, K.~Narasimhan, and T.~L. Griffiths.
\newblock Cognitive architectures for language agents.
\newblock {\em arXiv preprint arXiv:2309.02427}, 2023.

\bibitem{suzgun2022challenging}
M.~Suzgun, N.~Scales, N.~Sch{\"a}rli, S.~Gehrmann, Y.~Tay, H.~W. Chung, A.~Chowdhery, Q.~V. Le, E.~H. Chi, D.~Zhou, et~al.
\newblock Challenging big-bench tasks and whether chain-of-thought can solve them.
\newblock {\em arXiv preprint arXiv:2210.09261}, 2022.

\bibitem{2023internlm}
I.~Team.
\newblock Internlm: A multilingual language model with progressively enhanced capabilities.
\newblock \url{https://github.com/InternLM/InternLM-techreport}, 2023.

\bibitem{touvron2023llama}
H.~Touvron, T.~Lavril, G.~Izacard, X.~Martinet, M.-A. Lachaux, T.~Lacroix, B.~Rozière, N.~Goyal, E.~Hambro, F.~Azhar, A.~Rodriguez, A.~Joulin, E.~Grave, and G.~Lample.
\newblock Llama: Open and efficient foundation language models, 2023.

\bibitem{wei2022emergent}
J.~Wei, Y.~Tay, R.~Bommasani, C.~Raffel, B.~Zoph, S.~Borgeaud, D.~Yogatama, M.~Bosma, D.~Zhou, D.~Metzler, et~al.
\newblock Emergent abilities of large language models.
\newblock {\em arXiv preprint arXiv:2206.07682}, 2022.

\bibitem{wei2022chain}
J.~Wei, X.~Wang, D.~Schuurmans, M.~Bosma, F.~Xia, E.~Chi, Q.~V. Le, D.~Zhou, et~al.
\newblock Chain-of-thought prompting elicits reasoning in large language models.
\newblock {\em Advances in neural information processing systems}, 35:24824--24837, 2022.

\bibitem{yu2020reclor}
W.~Yu, Z.~Jiang, Y.~Dong, and J.~Feng.
\newblock Reclor: A reading comprehension dataset requiring logical reasoning.
\newblock {\em arXiv preprint arXiv:2002.04326}, 2020.

\bibitem{zeng2022glm}
A.~Zeng, X.~Liu, Z.~Du, Z.~Wang, H.~Lai, M.~Ding, Z.~Yang, Y.~Xu, W.~Zheng, X.~Xia, et~al.
\newblock Glm-130b: An open bilingual pre-trained model.
\newblock {\em arXiv preprint arXiv:2210.02414}, 2022.

\end{thebibliography}
\appendix
    \section{JSON prompt}
\label{sec:json prompt}
The evaluation JSON constrain prompt template we used to evaluate the performance of different models is shown in Figure ~\ref{fig:json-prompt}.

\begin{figure}[htbp]
\centering
\begin{promptbox}[JSON Prompt Template]
Please generate a JSON object that follows standard JSON formatting and indentation, containing a field named 'answer'. The 'answer' field should be a list of strings, where each string represents ... The 'process' field should be a list of strings, where each string ...

\textbf{Example: String Synthesis}

\textbf{Input:} 

Now there are four different types of blocks: [A], [B], [C], and {A}, which satisfy the following rules:
\begin{enumerate}
    \item One [A], one [B], and one [C] can be synthesized into one \{A\}
    \item One [A] and one [B] can be synthesized into one [C]
\end{enumerate}
Rule 1, Rule 2, Rule 1, Rule 2... continue cycling through these rules to synthesize until no further synthesis is possible using either rule.

Question: If we currently have four [A], four [B], and three [C], what will be the result after synthesis?

\textbf{Json Constraints:}

Please generate a JSON object that follows standard JSON formatting and indentation, containing a field named 'answer'. The 'answer' field should be a list of strings, where each element represents the number of different types of blocks, in the order of [A], [B], [C], {A}. For example, if there is 1 block of type [A], 0 blocks of type [B], 3 blocks of type [C], it should be represented as ["1", "0", "3", "0"]. The 'process' field should be a list of strings, where each string records the instructions for each step from the initial state to the final state. First output the blocks that need to be synthesized, followed by the "->" symbol, then output the synthesized block, without adding any extra explanations. For example: ["[A] [B] [C] -> \{A\}", "[A] [B] -> [C]"].

\textbf{Output:}
\begin{verbatim}
{
  "answer": ["0", "3", "6", "1"],
  "process": [
        "[A] [B] [C] -> {A}",
        "[A] [B] -> [C]",
        "[A] [B] [C] -> {A}",
        "[A] [B] -> [C]"
    ]
}
\end{verbatim}

\end{promptbox}
\caption{JSON prompt and an example}
\label{fig:json-prompt}
\end{figure}

\section{Data overview of different categories}
\label{sec:data_overview}
Table~\ref{tab:taxonomy} provides a detailed classification of categories along with the corresponding sample sizes as shown below.

\begin{table}[ht]
    \centering
    \small
    \resizebox{0.7\textwidth}{!}{
    \begin{tblr}{
        hline{1, Z} = {1pt, solid},
        hline{2-Y},
        cell{2}{1} = {r = 13}{m},
        cell{15}{1} = {r = 10}{m},
        vline{2},
        vline{3},
        vline{4},
        cell{25}{1} = {c = 3}{c},
        row{2-10} = {oneDColor},
        row{11-14} = {twoDColor},
        cell{2}{1} = {white},
        row{15-19} = {oneDColor},
        row{20-24} = {twoDColor},
        cell{15}{1} = {white}
    }
        \textbf{Categories} & \textbf{Basic Tasks}  & \textbf{Application} & \textbf{\#Samples} \\
         \textbf{Execution}  & Character search & & 12\\ 
         & String insertion & & 8\\
         & String synthesis & Synthesis and decomposition & 24\\
         & String deletion and modification & & 8\\
         & String rearrangement & & 8\\
         & String splitting & & 8\\
         & String processing & Mahjong-type & 16\\
         & Statistical counting\textsuperscript{$\lambda$} & & 8\\
         & New operator calculation\textsuperscript{$\lambda$} & & 12\\
         & Element operations & Lights out & 20\\  
         & Pattern recognition & Reversi & 16\\
         & Matrix transformation & 2048\textsuperscript{$\lambda$} & 24\\
         & Path movement & Pooling\textsuperscript{$\lambda$} & 16\\
         \textbf{Planning}& Single-choice self-reasoning & & 8\\ 
         & Constrained Linear Arrangement & & 16\\ 
         & Mutual generation and restriction & & 16\\ 
         & Logical equations\textsuperscript{$\lambda$} & & 8\\
         & Combinatorial calculation\textsuperscript{$\lambda$} & & 8\\
         & Eight Queens puzzle & Letter logic diagram & 16\\ 
         & Logic puzzle\textsuperscript{$\lambda$} & & 12\\
         & Minesweeper\textsuperscript{$\lambda$} & & 8\\
         & Standard Sudoku\textsuperscript{$\lambda$} & Sudoku with arithmetic rules\textsuperscript{$\lambda$} & 16\\
         & Cryptanalysis\textsuperscript{$\lambda$} & & 16\\
        \textbf{Total} & & & 304\\
    \end{tblr}
    }
    \vspace{1mm}
    \setlength{\belowcaptionskip}{-0.5em}%
    \caption{An overview of \model, including Execution and Planning. Light red rows indicate sequential-based tasks, light blue rows indicate matrix-based tasks. Tasks involving math are denoted with the superscript\textsuperscript{$\lambda$}}
    \label{tab:taxonomy}
\end{table}

\section{Additional evaluation metrics and error analysis}
\label{sec:error_metrics}
During the evaluation process, it is observed that some models occasionally fail to adhere to instructions regarding the constraint of JSON format output. In this context, we have defined two error metrics and one correctness metric for thorough analysis:
\begin{enumerate}
    \item \textbf{JSON Error (JSError):} This metric tracks instances where there is an error in parsing the JSON format, usually due to incomplete or improperly formatted JSON outputs.
     \item \textbf{Instruction Following Error (IFError):} This metric measures the number of instances in which a JSON format could not be successfully extracted.
     \item \textbf{Answer Process Accuracy based on non-IFError and non-JSError (NIJ-Acc):} This is the correctness metric, which evaluates the Answer Process Accuracy (AP-Acc) concerning non-IFError and non-JSError. The formula is provided in Equation~\ref{equ:CR}.
\end{enumerate}

\begin{equation}
    \label{equ:CR}
    S_{\mathrm{NIJ-Acc}} = \frac{S_{\mathrm{AP-Acc}}}{1- S_{\mathrm{IFError}} - S_{\mathrm{JSError}}}
\end{equation}

\begin{table}[t] \centering \resizebox{\linewidth}{!}{ \begin{tblr}{ hline{1, Z} = {1pt, solid}, hline{2} = {2-3}{leftpos = -1, rightpos = -1, endpos}, hline{2} = {4-5}{leftpos = -1, rightpos = -1, endpos}, hline{2} = {6-8}{leftpos = -1, rightpos = -1, endpos}, hline{3}, cell{1}{1} = {r = 2}{m}, cell{1}{2} = {c = 2}{c}, cell{1}{4} = {c = 2}{c}, cell{1}{6} = {c = 3}{c}, column{2-8} = {3-Z}{r}, row{1} = {c} } \textbf{Model} & \textbf{IFError(\%$\downarrow$)} & & \textbf{JSError(\%$\downarrow$)} & & \textbf{NIJ-Acc(\%$\uparrow$)} & & \\ & \textbf{Execution} & \textbf{Planning} & \textbf{Execution} & \textbf{Planning} & \textbf{Execution} & \textbf{Planning} & \textbf{Avg.}\\ 
\textbf{o1-preview} & 0.56 & 0.00 & 0.00 & 0.00 & \underline{57.54} & \underline{47.58} & \underline{53.47}\\
\textbf{o1-mini} & 0.56 & 0.81 & 0.56 & 0.00 & 53.37 & 45.53 & 50.17\\
\textbf{claude-3.5-sonnet} & 0.56 & 0.00 & 0.00 & 0.00 & 33.52 & 23.39 & 29.37\\ 
\textbf{gpt-4o} & 1.11 & 0.00 & 0.00 & 1.61 & 35.39 & 18.85 & 28.67\\ 
\textbf{gpt-4-turbo-0409} & 1.11 & 0.00 & 0.00 & \dag \textbf{2.42} & 22.47 & 19.83 & 21.40\\ 
\textbf{glm-4-plus} & 5.00 & 9.68 & 1.11 & 0.00 & 19.64 & 18.93 & 19.22\\ 
\textbf{qwen2-72b-instruct} & 30.00 & 16.94 & 0.56 & 0.81 & 12.00 & 11.76 & 11.89\\ 
\textbf{llama-3-70b-chat} & \dag \textbf{52.22} & \ddag \textbf{60.48} & 0.56 & 0.00 & 14.12 & 4.08 & 10.45 \\ 
\textbf{glm-4-9b} & 21.67 &  10.48 & 2.22 & 0.81 & 10.95 & 8.18 & 9.72\\
\textbf{internlm-2.5-7b-chat} & 15.56 & 9.68 & 2.22 & 0.00 & 9.46 & 4.46 & 7.31\\ 
\textbf{mistral-7b-instruct} &  19.44 & 14.52 &  \ddag \textbf{13.33} & \dag \textbf{2.42} & 8.26 & 2.91 & 5.80\\ 
\textbf{claude-3-haiku} & 0.00 & 0.00 & 1.11 & \dag\textbf{2.42} & 4.49 & 5.49 & 5.02\\ 
\textbf{qwen2-7b-instruct} & 2.22 & 0.81 &  \dag \textbf{11.67} & \ddag \textbf{3.23} & 1.94 & 1.68 & 1.82\\ 
\textbf{llama-3-8b-chat} &  \ddag \textbf{75.00} &  \dag \textbf{58.87} & 0.00 & 0.81 & 0.00 & 0.00 & 0.00\\ 
\end{tblr}
    }
    \vspace{1mm}
    \setlength{\belowcaptionskip}{-0.5em}%
    \caption{Model performance on JSError, ResNError, IFError and NIJ-Acc metrics of \textbf{\textit{en} version}, with the Avg. calculated as the arithmetic mean NIJ-Acc value of both execution and planning. \ddag and \dag shows the worst and second worst performance in error metrics respectively. \underline{Underline} shows the best performance in NIJ-Acc metric.}
    \label{tab:error_metrics_en}
\end{table}

\begin{table}[t] \centering \resizebox{\linewidth}{!}{ \begin{tblr}{ hline{1, Z} = {1pt, solid}, hline{2} = {2-3}{leftpos = -1, rightpos = -1, endpos}, hline{2} = {4-5}{leftpos = -1, rightpos = -1, endpos}, hline{2} = {6-8}{leftpos = -1, rightpos = -1, endpos}, hline{3}, cell{1}{1} = {r = 2}{m}, cell{1}{2} = {c = 2}{c}, cell{1}{4} = {c = 2}{c}, cell{1}{6} = {c = 3}{c}, column{2-8} = {3-Z}{r}, row{1} = {c} } \textbf{Model} & \textbf{IFError(\%$\downarrow$)} & & \textbf{JSError(\%$\downarrow$)} & & \textbf{NIJ-Acc(\%$\uparrow$)} & & \\ & \textbf{Execution} & \textbf{Planning} & \textbf{Execution} & \textbf{Planning} & \textbf{Execution} & \textbf{Planning} & \textbf{Avg.}\\ 
\textbf{o1-preview} & 0.00 & 0.00 & 0.00 & 0.00 & \underline{60.00} & \underline{47.58} & \underline{54.93}\\
\textbf{o1-mini} & 0.56 & 0.81 & 0.56 & 0.00 & 56.67 & 45.16 & 51.97\\
\textbf{claude-3.5-sonnet} & 0.00 & 0.00 & 0.00 & 0.00 & 36.67 & 20.97 & 30.26\\ 
\textbf{gpt-4o} & 0.56 & 0.00 & 0.00 & 0.00 & 30.17 & 22.58 & 27.06\\ 
\textbf{gpt-4-turbo-0409} & 0.56 & 1.61 & 0.00 & 1.61 & 25.70 & 26.67 & 26.09\\ 
\textbf{glm-4-plus} & 10.00 & 8.06 & 0.56 & 0.00 & 26.09 & 21.05 & 24.00\\ 
\textbf{qwen2-72b-instruct} & 3.33 & 1.61 & 0.56 & 1.61 & 23.70 & 17.50 & 21.16\\ 
\textbf{llama-3-70b-chat} & 12.78 & \dag \textbf{13.71} & 0.00 &  0.81 & 17.83 & 9.43 & 14.45 \\ 
\textbf{claude-3-haiku} & 3.89 & 0.81 & 0.00 & 0.81 & 9.83 & 9.84 & 9.83\\
\textbf{glm-4-9b} & 25.00 &  \dag \textbf{13.71} & 1.11 & 0.81 & 12.03 & 6.60 & 9.62\\
\textbf{llama-3-8b-chat} &  \ddag \textbf{40.56} &  \ddag \textbf{28.23} & 0.00 & 0.00 & 5.61 & 5.62 & 5.61\\ 
\textbf{internlm-2.5-7b-chat} & 17.78 & 1.61 & 1.67 & \ddag \textbf{5.65} & 5.52 & 5.22 & 5.38\\ 
\textbf{mistral-7b-instruct} &  \dag \textbf{36.67} & 11.29 &  \ddag \textbf{7.22} & \dag \textbf{4.84} & 1.98 & 6.73 & 4.39\\  
\textbf{qwen2-7b-instruct} & 3.89 & 2.42 &  \dag \textbf{2.22} & 4.03 & 1.18 & 5.17 & 2.81\\ 
\end{tblr}
    }
    \vspace{1mm}
    \setlength{\belowcaptionskip}{-0.5em}%
    \caption{Model performance on JSError, ResNError, IFError and NIJ-Acc metrics of \textbf{\textit{zh} version}, with the Avg. calculated as the arithmetic mean NIJ-Acc value of both execution and planning. \ddag and \dag shows the worst and second worst performance in error metrics respectively. \underline{Underline} shows the best performance in NIJ-Acc metric.}
    \label{tab:error_metrics_zh}
\end{table}

The metrics presented in Table~\ref{tab:error_metrics_en} and Table~\ref{tab:error_metrics_zh} provide a comprehensive assessment of the model's error frequency in \textit{en} and \textit{zh} version respectively. The NIJ-Acc metric evaluates the model's overall accuracy. In this study, the llama-3-8b-chat and llama-3-70b-chat model shows consistent poor instruction-following capabilities in both versions, while the mistral-7b-instruct model performs poorly in the metric of json format. Conversely, stronger models tend to perform more stably as observed in these metrics. Overall error probabilities (IFError, JSError) are low across most models, resulting in minimal variations in NIJ-Acc scores compared to the average scores in Table~\ref{tab:main_result}, with only minor rank adjustments among similarly performing models.

\section{Few shot results of \textit{en} version}
\label{sec:en_few_shot}
\begin{figure}
    \centering
    \includegraphics[width=\linewidth]{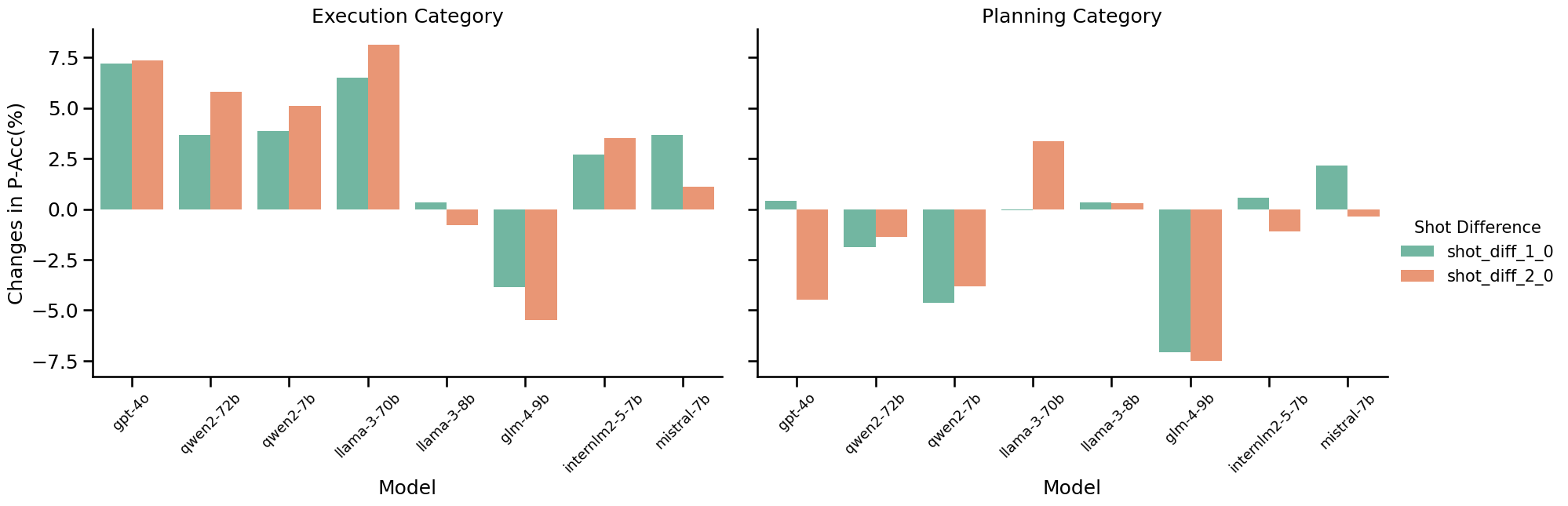}
    \caption{Few-shot differences on execution and planning category of \model's \textit{en} version. "shot\_diff\_1\_0" represents the difference in the P-Acc score between the 1-shot and 0-shot settings, calculated as the result of 1-shot minus the result of 0-shot, "shot\_diff\_2\_0" representing the P-Acc score between the 2-shot and 0-shot settings similarly.}
    \label{fig:category_diff_fewshot_en}
\end{figure}

\begin{figure}
    \centering
    \includegraphics[width=0.8\linewidth]{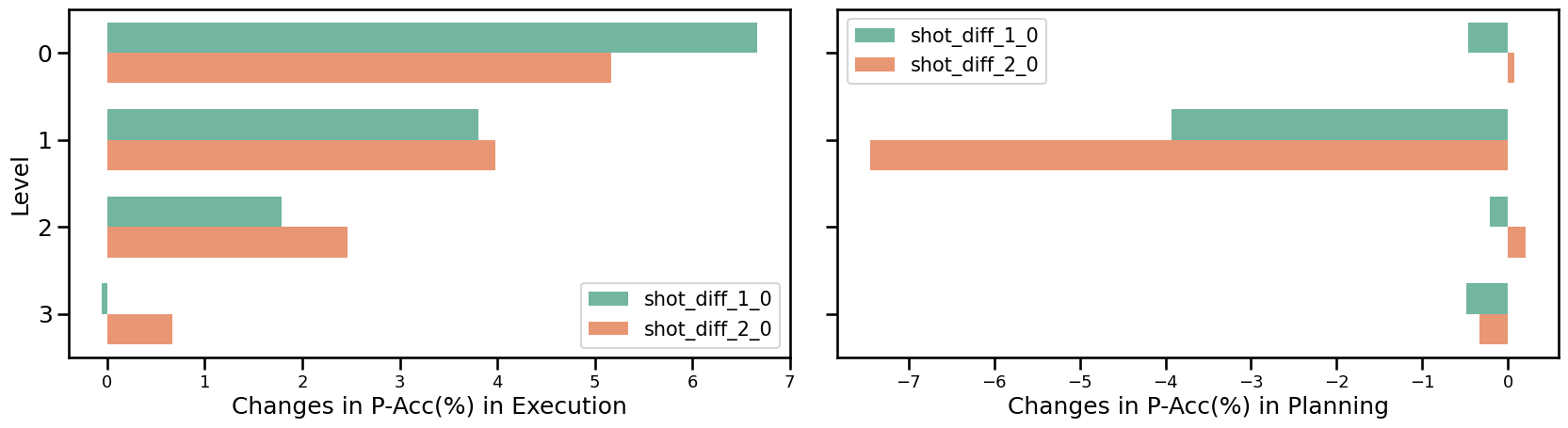}
    \caption{Few-shot differences on difficulty levels of \model's \textit{en} version with shot difference settings similar with Figure~\ref{fig:category_diff_fewshot_en}.}
    \label{fig:level_diff_fewshot_en}
\end{figure}

Figure~\ref{fig:category_diff_fewshot_en} and Figure~\ref{fig:level_diff_fewshot_en} shows few-shot results of \model's \textit{en} version. For 'Execution' category, the observed conclusion is aligned with \textit{zh} version. Shot contexts increases P-Acc score in most cases except glm-4-9b and llama-3-8b-chat. When difficulty levels are increased, the positive effect of shot contexts declines. For 'Planning' category, most  models decrease or has no change when extra shot contexts are added except for llama-3-70b-chat and mistral-7b-instruct. No notable effect on shot contexts on different difficulty levels except for level 1.

\section{Problems all models fail}
\label{sec:fail_heatmap}

\begin{figure}[h]
    \centering
    \includegraphics[width=\linewidth]{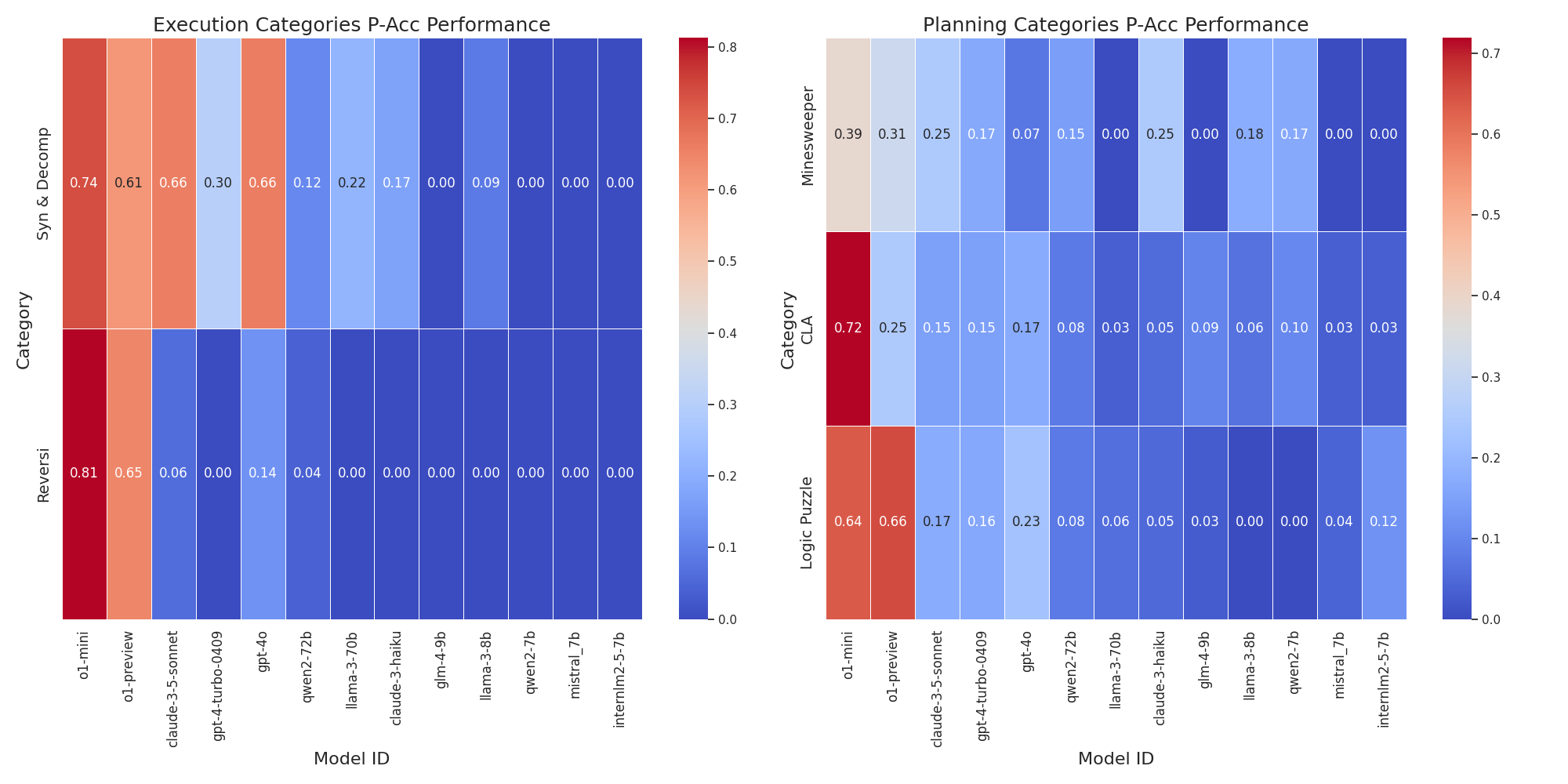}
    \caption{Average AP-Acc scores for five categories with the poorest performance: Reversi and Synthesis and Decomposition from the 'Execution' category, and Minesweeper, Constrained Linear Arrangement, and Logic Puzzle from the 'Planning' category.}
    \label{fig:wrong_heatmap}
\end{figure}

Figure~\ref{fig:wrong_heatmap} categorizes the five areas where all models exhibit the poorest performance, presenting the average AP-ACC scores for each category in a heat map. The horizontal axis corresponds to the model capabilities as outlined in Table~\ref{tab:main_result_en} and Table~\ref{tab:main_result_zh}. This figure highlights that models generally struggle most in two sub-categories within the execution scenario, particularly with 'Reversi', where many models score close to zero except for o1-mini and o1-preview models. Conversely, in planning scenarios such as 'Constrained Linear Arrangement', there is a slight variation in performance across different models. 

\end{document}